%% file: sn-article.tex
\documentclass[pdflatex,sn-nature]{sn-jnl} 

\usepackage{graphicx}%
\usepackage{multirow}%
\usepackage{amsmath,amssymb,amsfonts}%
\usepackage{amsthm}%
\usepackage{mathrsfs}%
\usepackage[title]{appendix}%
\usepackage{xcolor}%
\usepackage{textcomp}%
\usepackage{manyfoot}%
\usepackage{booktabs}%
\usepackage{algorithm}%
\usepackage{algorithmicx}%
\usepackage{algpseudocode}%
\usepackage{listings}%
\usepackage{placeins}%

\usepackage{booktabs}
\usepackage{longtable}
\usepackage{array}
\usepackage{hyperref}
\usepackage{caption}

\theoremstyle{thmstyleone}%
\theoremstyle{thmstyletwo}%

\theoremstyle{thmstylethree}%

\begin{document}

\title[Article Title]{Single Slice-to-3D Reconstruction in Medical Imaging and Natural Objects: A Comparative Benchmark with SAM 3D}


\author{\fnm{Yan} \sur{Luo}}\email{yluo16@meei.harvard.edu}
\equalcont{These authors contributed equally to this work.}

\author{\fnm{Advaith} \sur{Ravishankar}}\email{aravishankar@g.harvard.edu}
\equalcont{These authors contributed equally to this work.}

\author{\fnm{Serena} \sur{Liu}}\email{serenaliu1@college.harvard.edu}
\equalcont{These authors contributed equally to this work.}

\author{\fnm{Yutong} \sur{Yang}}\email{yutongyang@hsph.harvard.edu}

\author*{\fnm{Mengyu} \sur{Wang}}\email{mengyu\_wang@meei.harvard.edu}

\affil{\orgdiv{Harvard AI and Robotics Lab}, \orgname{Harvard University}}


\abstract{
While three-dimensional imaging is essential for clinical diagnosis, its high cost and long wait times have motivated the use of image-to-3D foundation models to infer volume from two-dimensional modalities. However, because these models are trained on natural images, their learned geometric priors struggle to transfer to inherently planar medical data. A benchmark of five state-of-the-art models (SAM3D, Hunyuan3D-2.1, Direct3D, Hi3DGen, and TripoSG) across six medical and two natural datasets revealed that voxel-based overlap remains uniformly low across all methods due to severe depth ambiguity from single-slice inputs. Despite this fundamental volumetric failure, global distance metrics indicate that SAM3D best captures topological similarity to ground-truth medical shapes, whereas alternative models are prone to oversimplification. Ultimately, these findings quantify the limits of zero-shot single-slice 3D inference, highlighting that reliable medical 3D reconstruction requires domain-specific adaptation and anatomical constraints to overcome complex medical geometries.
}



\maketitle
\pagebreak
\input{sec/intro}

\input{sec/results}

\input{sec/discussion}

\input{sec/method}

\section{Data Availability}
All datasets used in this work are publicly available. AeroPath is available at \url{https://github.com/raidionics/AeroPath}. BTCV is available at \url{https://www.kaggle.com/datasets/ipythonx/abdomen}. Duke C-Spine is available at \url{https://data.midrc.org/discovery/H6K0-A61V}. The Medical Segmentation Decathlon can be accessed at \url{http://medicaldecathlon.com/dataaws/}. Google Scanned Objects is available at \url{https://research.google/blog/scanned-objects-by-google-research-a-dataset-of-3d-scanned-common-household-items/}. Animal3D is available at \url{https://xujiacong.github.io/Animal3D/}.

\section{Code Availability}

The code used to reproduce the results presented in this work is publicly available at \url{https://github.com/luoyan407/Benchmark_3D_Generative_Models}.

\section{Acknowledgments}

This work was supported by NIH R00 EY028631, NIH R21 EY035298, NIH P30 EY003790, Research To Prevent Blindness International Research Collaborators Award, and Alcon Young Investigator Grant.

\section{Author Contributions}

Y.L. and M.W. conceived the study. Y.L. implemented the benchmark pipeline and conducted the experiments. A.R. and S.L. curated the data and assisted with some of the experiments. Formal analysis and visualization were performed by A.R., S.L., and Y.Y. M.W. was responsible for funding acquisition and project administration. The manuscript was drafted by Y.L., A.R., and S.L. All authors reviewed, edited, and approved the final version of the manuscript.

\section{Competing Interests}

The authors declare no competing interests.

\begin{appendices}
\section{Benchmark Performance Tables}\label{secA1}

\begin{table}[!h]
\centering
\makebox[\linewidth][c]{%
\renewcommand{\arraystretch}{1.15}
\begin{tabular}{llccccc}
\toprule
\textbf{Dataset} & \textbf{Model} &
\textbf{F1@0.01} & \textbf{Voxel-IoU} & \textbf{Voxel-Dice} & \textbf{CD} & \textbf{EMD} \\
\midrule

\multirow{5}{*}{AeroPath} & Hunyuan3D-2.1 & 0.0122 $\pm$ 0.0025 & 0.0200 $\pm$ 0.0042 & 0.0392 $\pm$ 0.0080 & 0.4200 $\pm$ 0.0344 & 0.4096 $\pm$ 0.0316 \\
                          & Direct3D      & 0.0118 $\pm$ 0.0022 & 0.0246 $\pm$ 0.0063 & 0.0479 $\pm$ 0.0120 & 0.3893 $\pm$ 0.0532 & 0.3882 $\pm$ 0.0454 \\
                          & Hi3DGen       & 0.0152 $\pm$ 0.0041 & 0.0385 $\pm$ 0.0083 & 0.0739 $\pm$ 0.0153 & 0.3036 $\pm$ 0.0501 & 0.3186 $\pm$ 0.0537 \\
                          & TripoSG       & 0.0141 $\pm$ 0.0030 & 0.0236 $\pm$ 0.0061 & 0.0461 $\pm$ 0.0117 & 0.4002 $\pm$ 0.0740 & 0.3961 $\pm$ 0.0719 \\
                          & SAM3D         & 0.0113 $\pm$ 0.0031 & 0.0223 $\pm$ 0.0079 & 0.0435 $\pm$ 0.0151 & 0.4075 $\pm$ 0.0527 & 0.3959 $\pm$ 0.0441 \\
\midrule

\multirow{5}{*}{Animal3D} & Hunyuan3D-2.1 & 0.0216 $\pm$ 0.0097 & 0.0435 $\pm$ 0.0171 & 0.0829 $\pm$ 0.0313 & 0.3601 $\pm$ 0.1329 & 0.3281 $\pm$ 0.1067 \\
                          & Direct3D      & 0.0186 $\pm$ 0.0094 & 0.0400 $\pm$ 0.0165 & 0.0764 $\pm$ 0.0297 & 0.3834 $\pm$ 0.0995 & 0.3649 $\pm$ 0.0779 \\
                          & Hi3DGen       & 0.0286 $\pm$ 0.0160 & 0.0605 $\pm$ 0.0288 & 0.1128 $\pm$ 0.0497 & 0.3134 $\pm$ 0.1147 & 0.3117 $\pm$ 0.0905 \\
                          & TripoSG       & 0.0335 $\pm$ 0.0169 & 0.0644 $\pm$ 0.0291 & 0.1196 $\pm$ 0.0504 & 0.2788 $\pm$ 0.0983 & 0.2861 $\pm$ 0.0873 \\
                          & SAM3D         & 0.0243 $\pm$ 0.0111 & 0.0486 $\pm$ 0.0180 & 0.0921 $\pm$ 0.0320 & 0.3717 $\pm$ 0.0940 & 0.3740 $\pm$ 0.0723 \\
\midrule

\multirow{5}{*}{BTCV}     & Hunyuan3D-2.1 & 0.0255 $\pm$ 0.0253 & 0.0359 $\pm$ 0.0294 & 0.0677 $\pm$ 0.0522 & 0.4140 $\pm$ 0.2643 & 0.3970 $\pm$ 0.1852 \\
                          & Direct3D      & 0.0429 $\pm$ 0.0437 & 0.0493 $\pm$ 0.0463 & 0.0905 $\pm$ 0.0780 & 0.3345 $\pm$ 0.1423 & 0.3502 $\pm$ 0.1288 \\
                          & Hi3DGen       & 0.0365 $\pm$ 0.0233 & 0.0595 $\pm$ 0.0334 & 0.1105 $\pm$ 0.0577 & 0.3340 $\pm$ 0.1389 & 0.3466 $\pm$ 0.1177 \\
                          & TripoSG       & 0.0443 $\pm$ 0.0434 & 0.0516 $\pm$ 0.0423 & 0.0953 $\pm$ 0.0720 & 0.3394 $\pm$ 0.1461 & 0.3615 $\pm$ 0.1372 \\
                          & SAM3D         & 0.0329 $\pm$ 0.0240 & 0.0494 $\pm$ 0.0344 & 0.0923 $\pm$ 0.0598 & 0.3356 $\pm$ 0.1380 & 0.3449 $\pm$ 0.1117 \\
\midrule

\multirow{5}{*}{Duke C-Spine} & Hunyuan3D-2.1 & 0.0687 $\pm$ 0.0403 & 0.0894 $\pm$ 0.0528 & 0.1599 $\pm$ 0.0871 & 0.2802 $\pm$ 0.2693 & 0.2872 $\pm$ 0.1793 \\
                             & Direct3D      & 0.0810 $\pm$ 0.0477 & 0.0978 $\pm$ 0.0519 & 0.1743 $\pm$ 0.0836 & 0.2360 $\pm$ 0.1859 & 0.2535 $\pm$ 0.1558 \\
                             & Hi3DGen       & 0.0771 $\pm$ 0.0667 & 0.1197 $\pm$ 0.1039 & 0.1989 $\pm$ 0.1608 & 0.4302 $\pm$ 0.5003 & 0.3665 $\pm$ 0.3331 \\
                             & TripoSG       & 0.0653 $\pm$ 0.0450 & 0.0892 $\pm$ 0.0658 & 0.1574 $\pm$ 0.1089 & 0.3874 $\pm$ 0.4332 & 0.3316 $\pm$ 0.2554 \\
                             & SAM3D         & 0.0955 $\pm$ 0.0648 & 0.1443 $\pm$ 0.0982 & 0.2397 $\pm$ 0.1464 & 0.1533 $\pm$ 0.1309 & 0.1766 $\pm$ 0.1116 \\
\midrule

\multirow{5}{*}{GSO}      & Hunyuan3D-2.1 & 0.0286 $\pm$ 0.0482 & 0.0596 $\pm$ 0.0731 & 0.1052 $\pm$ 0.1065 & 0.3980 $\pm$ 0.1809 & 0.3542 $\pm$ 0.1292 \\
                          & Direct3D      & 0.0290 $\pm$ 0.0392 & 0.0609 $\pm$ 0.0660 & 0.1088 $\pm$ 0.0976 & 0.3800 $\pm$ 0.1720 & 0.3409 $\pm$ 0.1226 \\
                          & Hi3DGen       & 0.0258 $\pm$ 0.0332 & 0.0600 $\pm$ 0.0670 & 0.1070 $\pm$ 0.0994 & 0.3868 $\pm$ 0.1752 & 0.3516 $\pm$ 0.1243 \\
                          & TripoSG       & 0.0302 $\pm$ 0.0539 & 0.0630 $\pm$ 0.0763 & 0.1109 $\pm$ 0.1068 & 0.3765 $\pm$ 0.1735 & 0.3401 $\pm$ 0.1233 \\
                          & SAM3D         & 0.0683 $\pm$ 0.0580 & 0.1814 $\pm$ 0.1363 & 0.2867 $\pm$ 0.1853 & 0.1547 $\pm$ 0.1783 & 0.1700 $\pm$ 0.1322 \\
\midrule

\multirow{5}{*}{MSD Brain} & Hunyuan3D-2.1 & 0.0105 $\pm$ 0.0116 & 0.0179 $\pm$ 0.0170 & 0.0345 $\pm$ 0.0318 & 0.5692 $\pm$ 0.2552 & 0.5611 $\pm$ 0.2297 \\
                           & Direct3D      & 0.0149 $\pm$ 0.0091 & 0.0211 $\pm$ 0.0142 & 0.0410 $\pm$ 0.0267 & 0.4211 $\pm$ 0.1170 & 0.4356 $\pm$ 0.1091 \\
                           & Hi3DGen       & 0.0197 $\pm$ 0.0108 & 0.0408 $\pm$ 0.0179 & 0.0778 $\pm$ 0.0326 & 0.3758 $\pm$ 0.1390 & 0.3840 $\pm$ 0.1259 \\
                           & TripoSG       & 0.0119 $\pm$ 0.0099 & 0.0185 $\pm$ 0.0189 & 0.0356 $\pm$ 0.0354 & 0.4598 $\pm$ 0.1437 & 0.4781 $\pm$ 0.1322 \\
                           & SAM3D         & 0.0158 $\pm$ 0.0081 & 0.0364 $\pm$ 0.0174 & 0.0697 $\pm$ 0.0318 & 0.3408 $\pm$ 0.1364 & 0.3505 $\pm$ 0.0988 \\
\midrule

\multirow{5}{*}{MSD Liver} & Hunyuan3D-2.1 & 0.0211 $\pm$ 0.0166 & 0.0322 $\pm$ 0.0214 & 0.0615 $\pm$ 0.0397 & 0.4483 $\pm$ 0.1970 & 0.4428 $\pm$ 0.1685 \\
                           & Direct3D      & 0.0264 $\pm$ 0.0173 & 0.0389 $\pm$ 0.0225 & 0.0740 $\pm$ 0.0413 & 0.4304 $\pm$ 0.1890 & 0.4249 $\pm$ 0.1612 \\
                           & Hi3DGen       & 0.0274 $\pm$ 0.0177 & 0.0488 $\pm$ 0.0264 & 0.0918 $\pm$ 0.0471 & 0.3959 $\pm$ 0.1994 & 0.4007 $\pm$ 0.1805 \\
                           & TripoSG       & 0.0260 $\pm$ 0.0221 & 0.0381 $\pm$ 0.0265 & 0.0722 $\pm$ 0.0472 & 0.3973 $\pm$ 0.1830 & 0.3960 $\pm$ 0.1461 \\
                           & SAM3D         & 0.0249 $\pm$ 0.0160 & 0.0419 $\pm$ 0.0232 & 0.0795 $\pm$ 0.0416 & 0.3813 $\pm$ 0.1510 & 0.3802 $\pm$ 0.1252 \\
\midrule

\multirow{5}{*}{MSD Lung}  & Hunyuan3D-2.1 & 0.0118 $\pm$ 0.0149 & 0.0178 $\pm$ 0.0212 & 0.0342 $\pm$ 0.0394 & 0.5803 $\pm$ 0.2084 & 0.5572 $\pm$ 0.1796 \\
                           & Direct3D      & 0.0135 $\pm$ 0.0113 & 0.0152 $\pm$ 0.0122 & 0.0297 $\pm$ 0.0233 & 0.5929 $\pm$ 0.1385 & 0.5745 $\pm$ 0.1193 \\
                           & Hi3DGen       & 0.0196 $\pm$ 0.0143 & 0.0404 $\pm$ 0.0223 & 0.0767 $\pm$ 0.0399 & 0.4220 $\pm$ 0.1201 & 0.4065 $\pm$ 0.1040 \\
                           & TripoSG       & 0.0128 $\pm$ 0.0214 & 0.0152 $\pm$ 0.0248 & 0.0289 $\pm$ 0.0435 & 0.5948 $\pm$ 0.1129 & 0.6014 $\pm$ 0.0930 \\
                           & SAM3D         & 0.0152 $\pm$ 0.0065 & 0.0351 $\pm$ 0.0154 & 0.0675 $\pm$ 0.0285 & 0.4240 $\pm$ 0.1205 & 0.4003 $\pm$ 0.0978 \\
                    
\bottomrule
\end{tabular}}
\caption{\textbf{Performance for coronal-view reconstructions.} Mean $\pm$ standard deviation of all five metrics across eight datasets. For F1, Voxel-IoU, and Voxel-Dice, higher is better; for CD and EMD, lower is better.}
\label{tab:benchmark_performance}
\end{table}

\begin{table}[!h]
\centering
\makebox[\linewidth][c]{%
\renewcommand{\arraystretch}{1.15}
\begin{tabular}{llccccc}
\toprule
\textbf{Dataset} & \textbf{Model} &
\textbf{F1@0.01} & \textbf{Voxel-IoU} & \textbf{Voxel-Dice} & \textbf{CD} & \textbf{EMD} \\
\midrule
\multirow{5}{*}{AeroPath} & Hunyuan3D-2.1 & 0.0102 $\pm$ 0.0025 & 0.0202 $\pm$ 0.0053 & 0.0395 $\pm$ 0.0102 & 0.4311 $\pm$ 0.0482 & 0.4382 $\pm$ 0.0377 \\
                          & Direct3D      & 0.0105 $\pm$ 0.0029 & 0.0208 $\pm$ 0.0060 & 0.0408 $\pm$ 0.0114 & 0.4306 $\pm$ 0.0480 & 0.4433 $\pm$ 0.0373 \\
                          & Hi3DGen       & 0.0134 $\pm$ 0.0025 & 0.0348 $\pm$ 0.0065 & 0.0672 $\pm$ 0.0122 & 0.3239 $\pm$ 0.0399 & 0.3538 $\pm$ 0.0422 \\
                          & TripoSG       & 0.0117 $\pm$ 0.0023 & 0.0257 $\pm$ 0.0054 & 0.0500 $\pm$ 0.0102 & 0.3586 $\pm$ 0.0887 & 0.3621 $\pm$ 0.0805 \\
                          & SAM3D         & 0.0117 $\pm$ 0.0038 & 0.0231 $\pm$ 0.0082 & 0.0451 $\pm$ 0.0156 & 0.4059 $\pm$ 0.0610 & 0.3981 $\pm$ 0.0591 \\
\midrule

\multirow{5}{*}{BTCV}     & Hunyuan3D-2.1 & 0.0184 $\pm$ 0.0228 & 0.0283 $\pm$ 0.0268 & 0.0539 $\pm$ 0.0486 & 0.5803 $\pm$ 0.3458 & 0.5143 $\pm$ 0.2422 \\
                          & Direct3D      & 0.0475 $\pm$ 0.0505 & 0.0576 $\pm$ 0.0566 & 0.1042 $\pm$ 0.0900 & 0.3157 $\pm$ 0.1299 & 0.3320 $\pm$ 0.1241 \\
                          & Hi3DGen       & 0.0301 $\pm$ 0.0222 & 0.0566 $\pm$ 0.0289 & 0.1057 $\pm$ 0.0509 & 0.3322 $\pm$ 0.1918 & 0.3371 $\pm$ 0.1365 \\
                          & TripoSG       & 0.0411 $\pm$ 0.0490 & 0.0514 $\pm$ 0.0571 & 0.0928 $\pm$ 0.0918 & 0.3577 $\pm$ 0.1853 & 0.3719 $\pm$ 0.1452 \\
                          & SAM3D         & 0.0280 $\pm$ 0.0206 & 0.0510 $\pm$ 0.0261 & 0.0960 $\pm$ 0.0463 & 0.3002 $\pm$ 0.1083 & 0.3109 $\pm$ 0.0850 \\
\midrule

\multirow{5}{*}{Duke C-Spine} & Hunyuan3D-2.1 & 0.0625 $\pm$ 0.0334 & 0.0709 $\pm$ 0.0367 & 0.1303 $\pm$ 0.0629 & 0.2725 $\pm$ 0.2848 & 0.2747 $\pm$ 0.1820 \\
                             & Direct3D      & 0.0807 $\pm$ 0.0327 & 0.0881 $\pm$ 0.0350 & 0.1601 $\pm$ 0.0577 & 0.1404 $\pm$ 0.0633 & 0.1772 $\pm$ 0.0689 \\
                             & Hi3DGen       & 0.0810 $\pm$ 0.0676 & 0.1140 $\pm$ 0.0939 & 0.1925 $\pm$ 0.1444 & 0.3883 $\pm$ 0.4225 & 0.3493 $\pm$ 0.2773 \\
                             & TripoSG       & 0.0876 $\pm$ 0.0539 & 0.1111 $\pm$ 0.0683 & 0.1933 $\pm$ 0.1089 & 0.3066 $\pm$ 0.3883 & 0.2862 $\pm$ 0.2272 \\
                             & SAM3D         & 0.0779 $\pm$ 0.0575 & 0.0966 $\pm$ 0.0799 & 0.1672 $\pm$ 0.1239 & 0.2092 $\pm$ 0.2142 & 0.2208 $\pm$ 0.1643 \\
\midrule

\multirow{5}{*}{MSD Brain} & Hunyuan3D-2.1 & 0.0122 $\pm$ 0.0131 & 0.0201 $\pm$ 0.0185 & 0.0388 $\pm$ 0.0344 & 0.5205 $\pm$ 0.2198 & 0.5183 $\pm$ 0.1966 \\
                           & Direct3D      & 0.0163 $\pm$ 0.0103 & 0.0239 $\pm$ 0.0149 & 0.0463 $\pm$ 0.0279 & 0.4085 $\pm$ 0.1163 & 0.4251 $\pm$ 0.1077 \\
                           & Hi3DGen       & 0.0181 $\pm$ 0.0107 & 0.0374 $\pm$ 0.0164 & 0.0716 $\pm$ 0.0299 & 0.3864 $\pm$ 0.1309 & 0.3871 $\pm$ 0.1150 \\
                           & TripoSG       & 0.0122 $\pm$ 0.0093 & 0.0189 $\pm$ 0.0181 & 0.0365 $\pm$ 0.0339 & 0.4568 $\pm$ 0.1400 & 0.4731 $\pm$ 0.1258 \\
                           & SAM3D         & 0.0146 $\pm$ 0.0061 & 0.0335 $\pm$ 0.0136 & 0.0644 $\pm$ 0.0253 & 0.3200 $\pm$ 0.0751 & 0.3292 $\pm$ 0.0717 \\
\midrule

\multirow{5}{*}{MSD Liver} & Hunyuan3D-2.1 & 0.0227 $\pm$ 0.0204 & 0.0350 $\pm$ 0.0233 & 0.0666 $\pm$ 0.0430 & 0.4802 $\pm$ 0.2556 & 0.4735 $\pm$ 0.2246 \\
                           & Direct3D      & 0.0234 $\pm$ 0.0146 & 0.0385 $\pm$ 0.0238 & 0.0731 $\pm$ 0.0432 & 0.4498 $\pm$ 0.2389 & 0.4385 $\pm$ 0.1899 \\
                           & Hi3DGen       & 0.0220 $\pm$ 0.0144 & 0.0452 $\pm$ 0.0280 & 0.0851 $\pm$ 0.0492 & 0.3919 $\pm$ 0.2443 & 0.3765 $\pm$ 0.1862 \\
                           & TripoSG       & 0.0243 $\pm$ 0.0168 & 0.0378 $\pm$ 0.0234 & 0.0718 $\pm$ 0.0429 & 0.4079 $\pm$ 0.1888 & 0.4090 $\pm$ 0.1557 \\
                           & SAM3D         & 0.0268 $\pm$ 0.0099 & 0.0497 $\pm$ 0.0238 & 0.0938 $\pm$ 0.0428 & 0.3275 $\pm$ 0.0990 & 0.3413 $\pm$ 0.1026 \\
\midrule

\multirow{5}{*}{MSD Lung}  & Hunyuan3D-2.1 & 0.0146 $\pm$ 0.0288 & 0.0203 $\pm$ 0.0257 & 0.0387 $\pm$ 0.0445 & 0.6005 $\pm$ 0.2194 & 0.5653 $\pm$ 0.1992 \\
                           & Direct3D      & 0.0138 $\pm$ 0.0119 & 0.0171 $\pm$ 0.0147 & 0.0332 $\pm$ 0.0277 & 0.5810 $\pm$ 0.1417 & 0.5563 $\pm$ 0.1227 \\
                           & Hi3DGen       & 0.0154 $\pm$ 0.0083 & 0.0367 $\pm$ 0.0169 & 0.0703 $\pm$ 0.0312 & 0.3874 $\pm$ 0.1335 & 0.3620 $\pm$ 0.1094 \\
                           & TripoSG       & 0.0107 $\pm$ 0.0149 & 0.0132 $\pm$ 0.0174 & 0.0254 $\pm$ 0.0323 & 0.6391 $\pm$ 0.1985 & 0.6114 $\pm$ 0.1411 \\
                           & SAM3D         & 0.0166 $\pm$ 0.0098 & 0.0382 $\pm$ 0.0178 & 0.0730 $\pm$ 0.0321 & 0.3657 $\pm$ 0.1014 & 0.3475 $\pm$ 0.0808 \\

\bottomrule
\end{tabular}}
\caption{
\textbf{Benchmark performance for axial-view reconstructions.} Mean $\pm$ standard deviation of all five metrics across six medical datasets. For F1, Voxel-IoU, and Voxel-Dice, higher is better; for CD and EMD, lower is better.
}
\label{tab:benchmark_performance}
\end{table}

\section{Additional Medical Samples}\label{secB}

\begin{figure}[!t]
\centering
\makebox[\linewidth][c]{%
\includegraphics[width=1.5\textwidth]{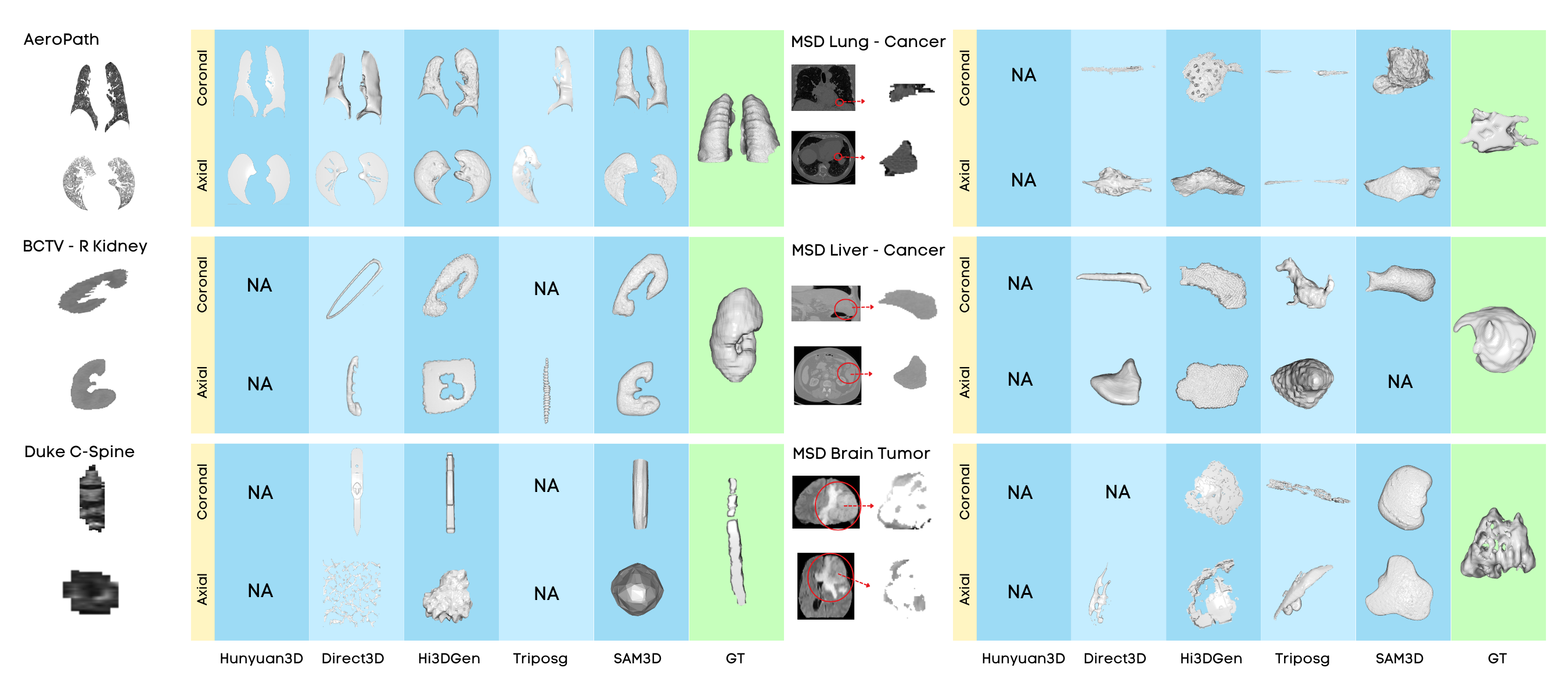}}
\caption{
\textbf{Additional qualitative medical samples across all datasets.} Coronal and axial reconstructions are shown for each method. MSD Brain tumor scans are T1-weighted with gadolinium contrast (T1Gd).
}
\end{figure}

\section{Additional Natural Samples}\label{secC}
\begin{figure}[!t]
\centering
\makebox[\linewidth][c]{%
\includegraphics[width=1.5\textwidth]{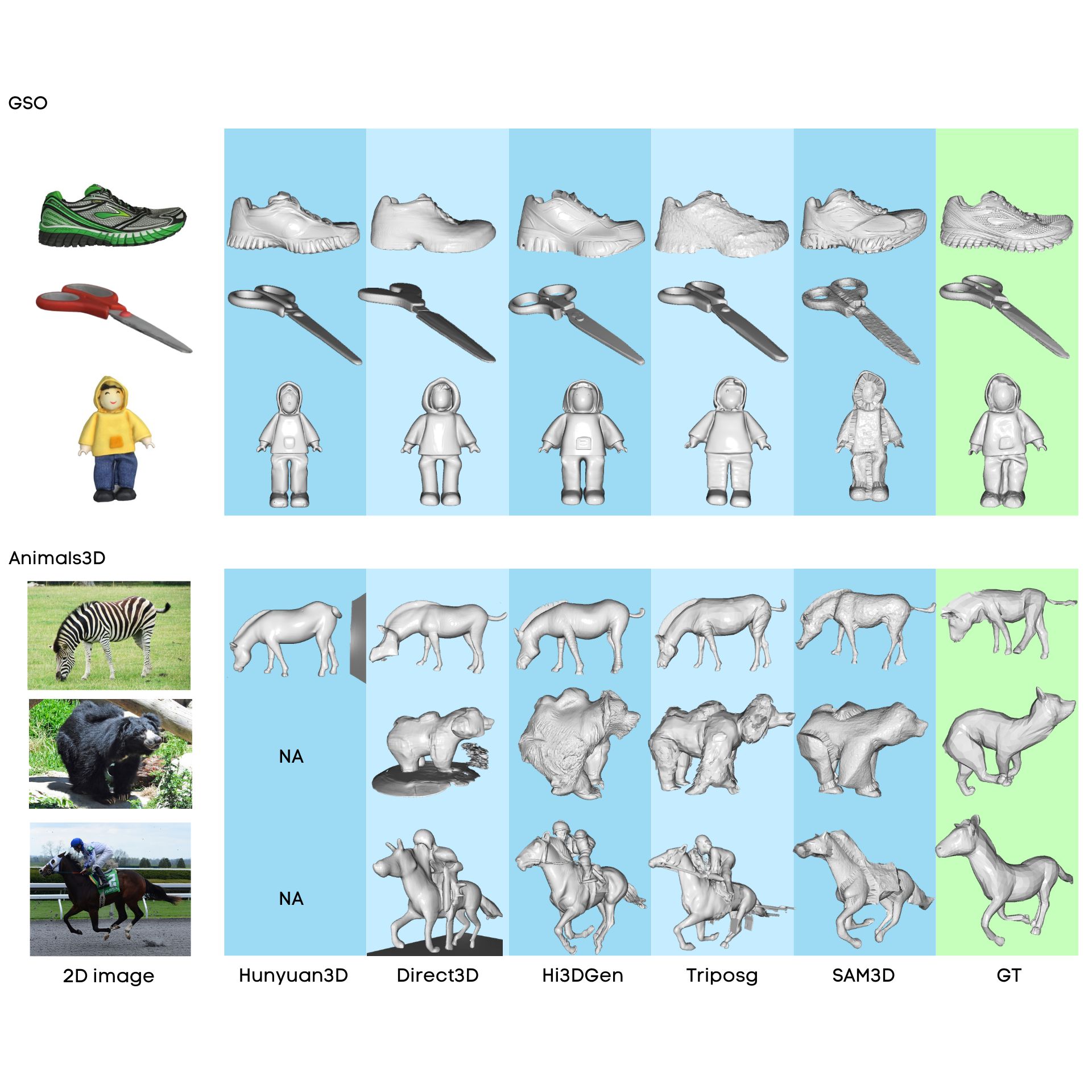}}
\caption{
\textbf{Additional qualitative natural samples.} Reconstructions from all five models on Animal3D and GSO, compared against ground truth.
}
\end{figure}




\end{appendices}

\pagebreak

\bibliography{sn-bibliography}

\end{document}

%% file: sec/intro.tex
\section{Introduction}\label{sec:introduction}
Three-dimensional representation of anatomical and pathological structures is central to clinical practice, such as clinical diagnosis and treatment planning in CT and MRI. Accurate assessment requires spatial relationships, volumetric measurements, and morphological features that two-dimensional imaging cannot adequately capture. The limitations of 2D imaging manifest across multiple clinical domains. In cardiovascular imaging, 2D echocardiography can underestimate ventricular volume compared to 3D methods, potentially delaying diagnosis \cite{Kawamura2014}. In breast imaging, digital breast tomosynthesis (DBT) reveals abnormalities masked by overlapping tissues in 2D mammography \cite{dbt_detection_advantage,Dhamija_Gulati_Deo_Gogia_Hari_2021}. Similarly, chest CT visualizes nodules and lesions that may be missed on 2D radiographs \cite{Gefter_Hatabu_2023}. Despite these benefits, 3D imaging techniques remain costly with long acquisition times \cite{Foti_Longo_2024,GharehMohammadi_Sebro_2024}.

Image-to-3D reconstruction methods address these limitations by inferring volumetric structures from readily available 2D imaging data. Initial generative approaches, such as X2CT-GAN \cite{ying2019x2ctganreconstructingctbiplanar}, demonstrated 3D reconstruction from limited radiographic views. Subsequent methods improved upon this with single-input approaches. Sli2Vol \cite{yeung2021sli2vol} achieved self-supervised slice-to-slice propagation within volumetric scans, while X-diffusion \cite{bourigault2025xdiffusiongeneratingdetailed3d} generated MRI volumes from single cross-sectional slices using diffusion models. However, these methods have significant limitations: X2CT-GAN requires multiple input views, X-diffusion is modality-specific, and Sli2Vol requires access to the full 3D volume during training, precluding true single-view reconstruction from arbitrary 2D slices. This motivates approaches that can recover geometry and texture from a single 2D image and generalize to novel anatomical and pathological structures without task-specific training.

Recent image-to-3D foundation models trained on large-scale natural image datasets offer a different paradigm: they learn universal geometric priors that generalize across domains without task-specific training. Models such as Hunyuan3D-2.1 \cite{hunyuan3d2025hunyuan3d21}, Direct3D \cite{wu2024direct3d}, TripoSG \cite{li2025triposg}, Hi3DGen \cite{ye2025hi3dgen}, and SAM3D \cite{sam3dteam2025sam3d3dfyimages} demonstrate strong reconstruction of naturalistic objects by inferring 3D geometry from single 2D images. Unlike medical-specific methods tailored to particular modalities or requiring complete volumetric scans, these foundation models enable zero-shot single-view reconstruction across different imaging modalities and diverse anatomical and pathological structures. However, a critical open question remains: do the geometric priors learned from natural images, where depth cues arise from shading, occlusion, and multi-object spatial relationships, transfer to medical imaging, where slices are inherently planar with minimal texture and uniform depth? Understanding this transferability is essential: if foundation models can achieve reasonable zero-shot medical reconstruction, they offer a scalable solution to the data scarcity, annotation costs, and modality-specific constraints that limit current medical-only approaches. Conversely, identifying their failure modes reveals fundamental domain gaps requiring targeted adaptation strategies.

\begin{figure}[!t]
\centering
\makebox[\linewidth][c]{%
\includegraphics[width=1.4\textwidth]{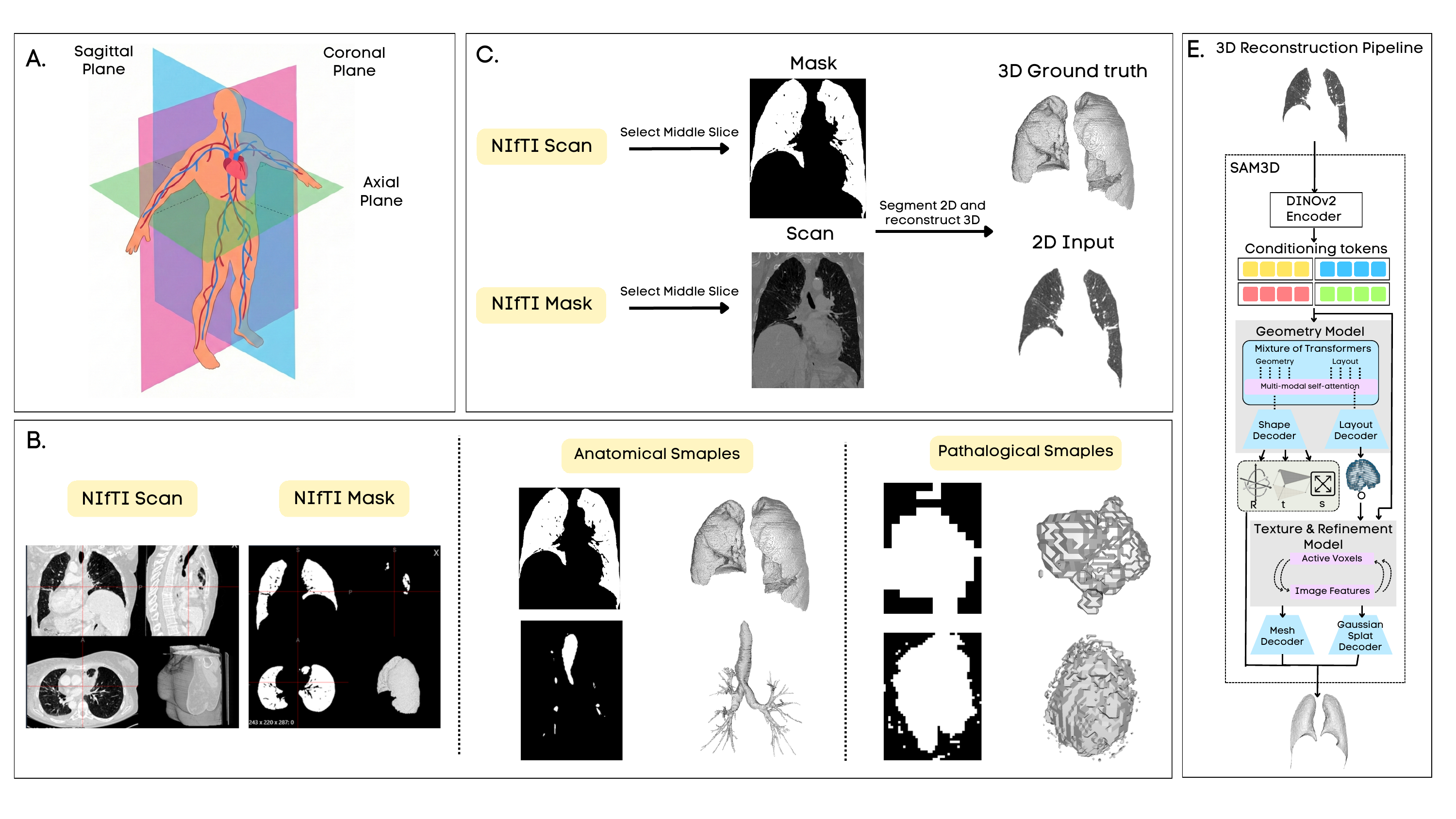}}
\caption{
\textbf{Zero-shot medical single-slice 3D reconstruction pipeline.} (A)~Three orthogonal planes (sagittal, coronal, axial) of a NIfTI volume. (B)~Example anatomical and pathological scan-mask pairs. (C)~A midpoint slice and its binary mask are extracted from the NIfTI scan and segmentation volume, respectively; the mask is applied to the slice to produce a 2D input image, while surface points extracted from the 3D mask serve as ground truth. (D)~The masked 2D input is passed through each reconstruction model to generate a predicted 3D point cloud, which is evaluated against the ground-truth point cloud using the metrics defined in Section~\ref{metrics}.
}
\label{fig:SAM3D_fig}
\end{figure}

We benchmark five state-of-the-art image-to-3D foundation models (SAM3D \cite{sam3dteam2025sam3d3dfyimages}, Hunyuan3D-2.1 \cite{hunyuan3d2025hunyuan3d21}, Direct3D \cite{wu2024direct3d}, Hi3DGen \cite{ye2025hi3dgen}, and TripoSG \cite{li2025triposg}) on six medical datasets containing anatomical and pathological structures in CT and MRI, following the zero-shot pipeline illustrated in Figure~\ref{fig:SAM3D_fig}. As shown in Figure~\ref{fig:SAM3D_fig}A–B, each dataset provides volumetric NIfTI scans with corresponding segmentation masks across both anatomical and pathological structures. A midpoint slice is extracted and masked to produce a single 2D input (Figure~\ref{fig:SAM3D_fig}C), which is then passed through each reconstruction model to generate a predicted 3D point cloud evaluated against the ground-truth surface geometry (Figure~\ref{fig:SAM3D_fig}D). Our zero-shot evaluation reveals a fundamental limitation: 2D medical slices lack the shading gradients, occlusion boundaries, and multi-object depth cues of natural images, leaving depth inference severely underconstrained. Consequently, all methods produce near-planar reconstructions with minimal volumetric extent, resulting in uniformly low voxel overlap. Despite these volumetric failures, global distance metrics indicate that SAM3D better captures overall shape distribution and coarse morphological features than alternative models. Furthermore, irregular, non-convex pathological structures compound this difficulty, yielding significantly worse reconstructions than simpler anatomical targets. Our findings quantify the limits of zero-shot single-slice medical reconstruction, emphasizing that reliable medical 3D reconstruction demands domain-specific adaptation and anatomical priors to address complex medical geometries.

%% file: sec/results.tex
\section{Results}\label{sec:result}


\subsection{Voxel-Based Reconstruction Quality}
\begin{figure}[!t]
    \centering
    \makebox[\linewidth][c]{%
    \includegraphics[width=1.5\textwidth]{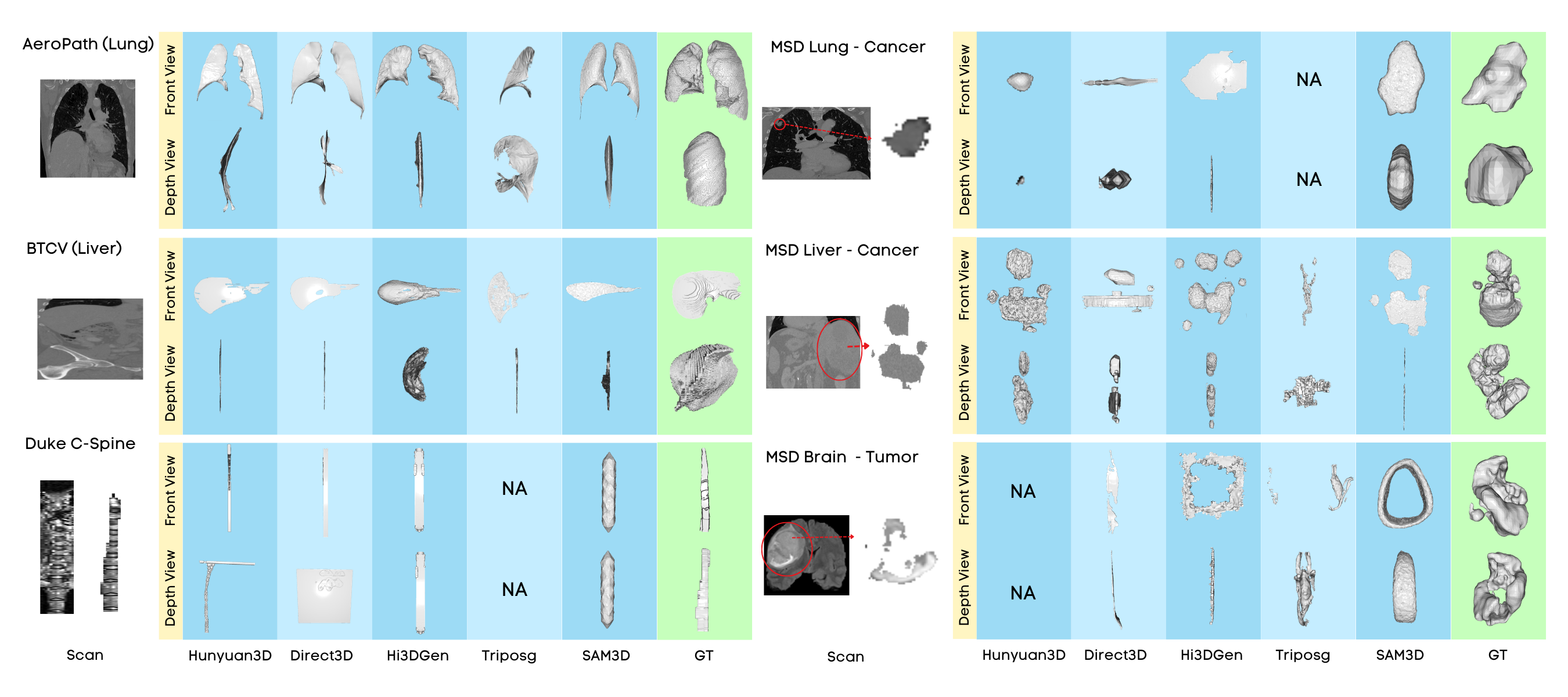}
    }
    \caption{
    \textbf{Coronal-view qualitative evaluation across medical datasets.} Each row shows the input scan followed by frontal and depth renderings of the reconstructed geometry from each method, compared against the ground truth (GT). MSD Brain Tumor scans are T2-weighted. Additional examples are provided in Appendix~\ref{secB}.
    }
    \label{fig:qualitative_results}
\end{figure}

\begin{figure}[!t]
    \centering
    \makebox[\linewidth][c]{%
    \includegraphics[width=1.2\textwidth]{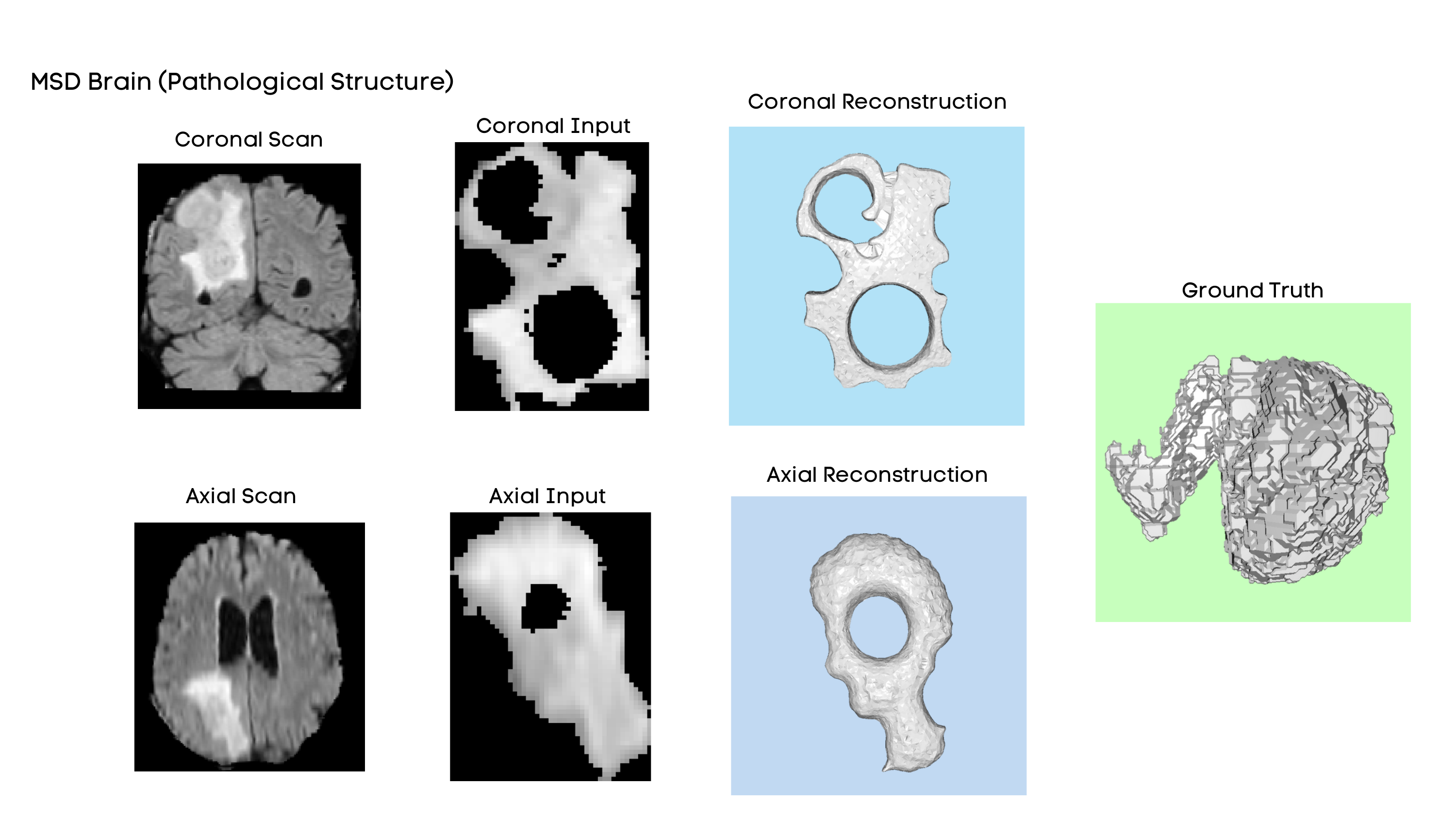}
    }
    \caption{
    \textbf{Coronal versus axial reconstruction of an MSD Brain tumor by SAM3D.} The coronal and axial input scans (left), their corresponding masked inputs (center-left), and SAM3D reconstructions (center-right) are shown alongside the ground-truth 3D surface (right).
    }
    \label{fig:coronal_vs_axial}
\end{figure}

\begin{figure}[!t]
\centering
\makebox[\linewidth][c]{%
\includegraphics[width=1.5\textwidth]{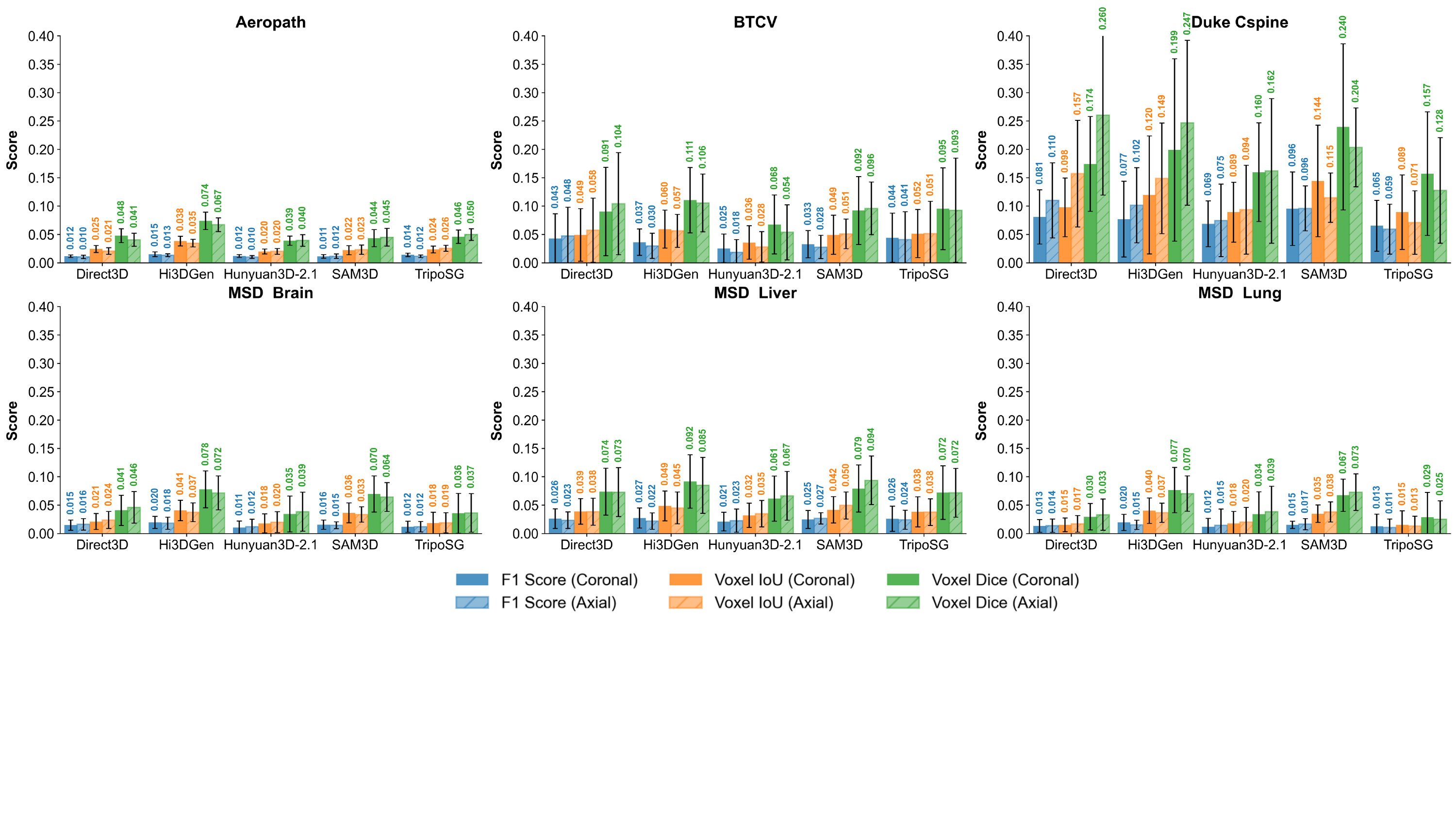}}
\caption{
\textbf{Voxel-based reconstruction quality across medical datasets.} F1 Score, Voxel IoU, and Voxel Dice for five models (Direct3D, Hi3DGen, Hunyuan3D-2.1, SAM3D, TripoSG) on six medical datasets. Solid bars denote coronal and hatched bars denote axial slice reconstructions. Error bars indicate standard deviation.
}
\label{fig:voxel_metrics}
\end{figure}

Figure \ref{fig:voxel_metrics} summarizes voxel-based reconstruction quality across all six medical datasets for coronal and axial slice inputs. All five models exhibited uniformly low voxel-based scores, with F1 scores largely below 0.10, Voxel IoU below 0.16, and Voxel Dice below 0.26 across datasets. Qualitative examples of these reconstructions are shown in Figure~\ref{fig:qualitative_results}. Performance also varied by input plane: coronal and axial reconstructions of the same model and dataset frequently diverged, indicating that reconstruction quality is plane-dependent. Figure~\ref{fig:coronal_vs_axial} illustrates this divergence for SAM3D \cite{sam3dteam2025sam3d3dfyimages} on an MSD Brain tumor, where coronal and axial inputs yield visually distinct reconstructions of the same structure. Additionally, pathological datasets (MSD Lung, MSD Brain, MSD Liver \cite{antonelli2022medical}) generally yielded lower voxel overlap scores than anatomical datasets (AeroPath \cite{stoverud2024aeropath}, BTCV,  \cite{zhou2025duke}), with the gap most pronounced for structures with highly irregular or non-convex morphologies.

Among the five benchmarked models, SAM3D \cite{chen2025sam,sam3dteam2025sam3d3dfyimages} and Hi3DGen \cite{ye2025hi3dgen} tended to achieve the highest voxel-based scores, though the margins of improvement were small and inconsistent across datasets. On MSD Lung, SAM3D and Hi3DGen reached Voxel IoU values of 0.038 and 0.039, respectively, compared to 0.019, 0.015, and 0.014 for Hunyuan3D-2.1 \cite{hunyuan3d2025hunyuan3d}, Direct3D \cite{wu2024direct3d}, and TripoSG \cite{li2025triposg}. A similar separation was observed on MSD Brain, where SAM3D and Hi3DGen achieved Voxel IoU of 0.034 and 0.039, approximately doubling the scores of the remaining models. Conversely, TripoSG and Hunyuan3D-2.1 frequently produced near-zero F1 scores on these datasets, indicating that they sometimes failed to generate any volumetrically meaningful overlap with the ground-truth anatomy. Even the best-performing model on any given dataset rarely exceeded a Voxel Dice of 0.25 or Voxel IoU of 0.15, upper bounds observed only on the comparatively simpler Duke C-Spine dataset.

\subsection{Point Cloud Distance Metrics}

\begin{figure}[!t]
\centering
\makebox[\linewidth][c]{%
\includegraphics[width=1.5\textwidth]{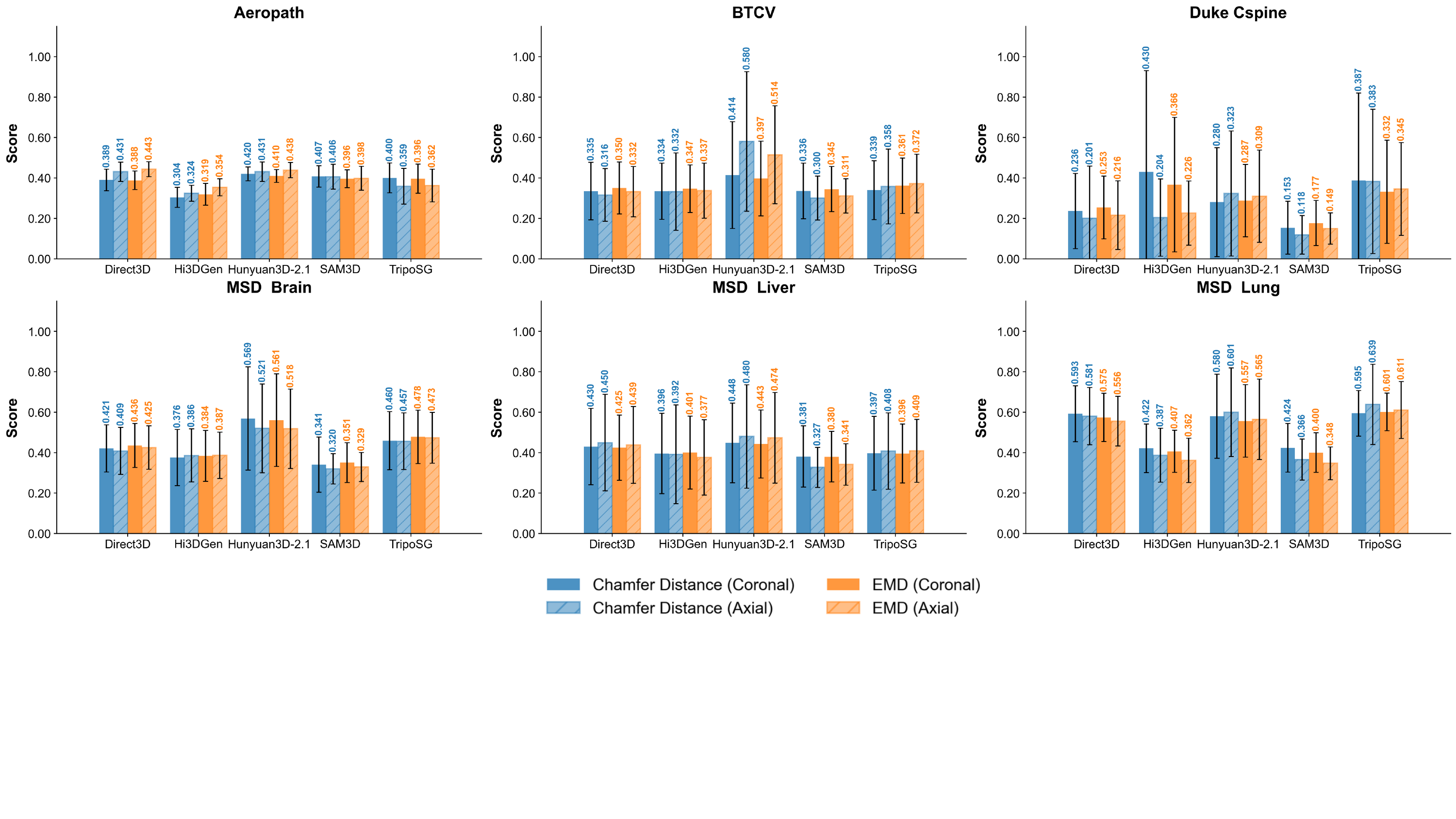}}
\caption{
\textbf{Point cloud distance metrics across medical datasets.} Chamfer Distance (CD) and Earth Mover's Distance (EMD) for five models (Direct3D, Hi3DGen, Hunyuan3D-2.1, SAM3D, TripoSG) on six medical datasets. Solid bars denote coronal and hatched bars denote axial slice reconstructions; lower values indicate better reconstruction quality. Error bars indicate standard deviation.
}
\label{fig:chamfer_emd_metrics}
\end{figure}

Figure \ref{fig:chamfer_emd_metrics} summarizes point cloud distance metrics across the six medical datasets. SAM3D consistently achieved the lowest or near-lowest Chamfer Distance (CD) and Earth Mover's Distance (EMD) across all medical datasets, indicating that its reconstructed point clouds most closely approximated the ground-truth shape distributions. This advantage was particularly pronounced on Duke C-Spine, where SAM3D's CD (0.153) and EMD (0.177) were roughly half those of the next best competitor, Direct3D (CD: 0.236; EMD: 0.254). In contrast to the voxel-based metrics in Figure \ref{fig:voxel_metrics}, which were uniformly poor across all methods, CD and EMD revealed clearer performance hierarchies among the five models, providing more meaningful separation between approaches. As with the voxel results, coronal and axial reconstructions continued to diverge, with certain plane-dataset combinations yielding substantially different distance values for the same model. Pathological datasets (e.g., MSD Lung, MSD Brain) also tended to produce higher CD and EMD than anatomical datasets, with MSD Lung reaching CD values above 0.55 for several models.

\subsection{Voxel Reconstruction on Natural Datasets}
\begin{figure}[!t]
\centering
\makebox[\linewidth][c]{%
\includegraphics[width=1.5\textwidth]{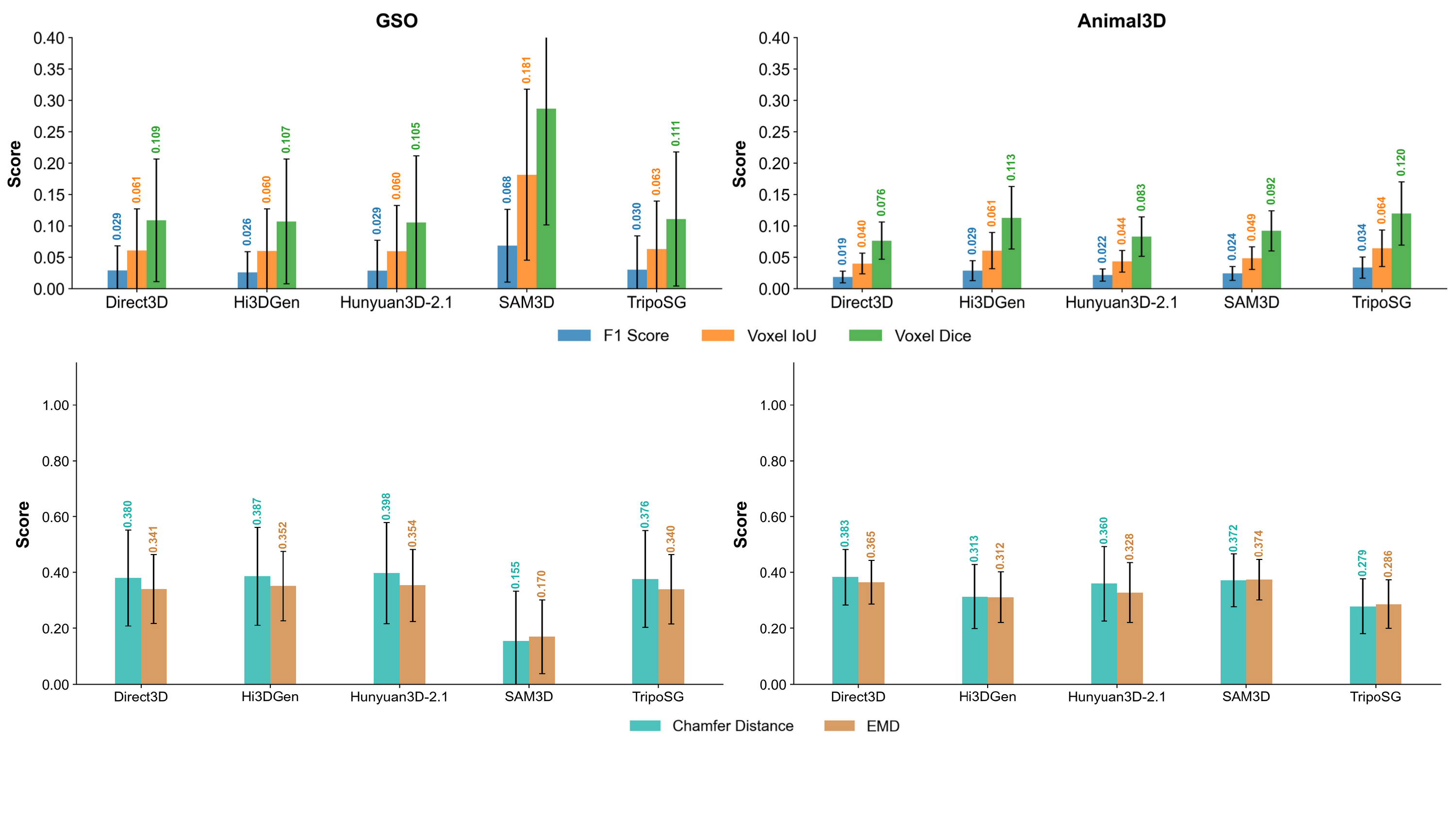}}
\caption{
\textbf{Reconstruction quality on natural datasets.} F1 Score, Voxel IoU, Voxel Dice, Chamfer Distance, and EMD for five models (Direct3D, Hi3DGen, Hunyuan3D-2.1, SAM3D, TripoSG) on Google Scanned Objects (GSO) and Animal3D.
}
\label{fig:natural_metrics}
\end{figure}

Figure \ref{fig:natural_metrics} reports voxel and distance-based metrics for the two natural datasets, Google Scanned Objects (GSO) \cite{downs2022google} and Animal3D \cite{xu2023animal3d}. Natural datasets achieved substantially higher voxel-based scores than medical datasets across all models: on GSO, SAM3D reached a Voxel IoU of approximately 0.18 and Voxel Dice near 0.29, far exceeding the best medical dataset results observed on Duke C-Spine (~0.14 IoU and ~0.24 Dice). SAM3D dominated on GSO across all metrics, achieving markedly higher F1, Voxel IoU, and Voxel Dice than all competitors while also producing the lowest CD (~0.155) and EMD (~0.170); the remaining four models clustered together with similar, substantially weaker performance. However, on Animal3D, the competitive landscape shifted: TripoSG and Hi3DGen led on voxel metrics, with TripoSG achieving the highest F1 and Voxel IoU, while SAM3D performed more moderately. Distance metrics on both natural datasets were also generally lower (better) than on medical datasets, with GSO and Animal3D yielding CD and EMD values in the 0.15–0.40 range compared to the 0.30–0.60 range observed across most medical datasets.

\subsection{Anatomical Versus Pathological Reconstruction Performance}

Across the six medical datasets, pathological datasets consistently yielded the weakest reconstruction performance. MSD Lung, MSD Brain, and MSD Liver produced lower voxel-based scores and higher distance metrics than the anatomical datasets (AeroPath, BTCV, Duke C-Spine) across all five models. On MSD Lung, for example, Voxel IoU values ranged from 0.014 (TripoSG) to 0.039 (Hi3DGen), and CD values exceeded 0.36 even for the best-performing model (SAM3D: 0.366), whereas the anatomical Duke C-Spine dataset achieved Voxel IoU up to 0.144 (SAM3D) and CD as low as 0.153 (SAM3D). MSD Brain exhibited a similar pattern, with all models producing Voxel Dice below 0.075 and CD values remaining above 0.319. MSD Liver fell between the two extremes but still underperformed relative to anatomical datasets, with the highest Voxel Dice reaching only 0.092 (SAM3D) and CD values remaining above 0.327 across all models.

\subsection{Cross-Dataset Performance Profile}
As shown in Figure \ref{fig:dataset_profile}, natural datasets generally outperformed medical datasets on distance metrics, with GSO (SAM3D CD: 0.155) and Animal3D (TripoSG CD: 0.279) achieving lower distance values than most medical datasets, with Duke C-Spine as the notable exception. Among the six medical datasets, Duke C-Spine yielded the highest absolute voxel-based scores: SAM3D reached a peak medical Voxel IoU of 0.144, Voxel Dice of 0.240, and F1 of 0.096, substantially exceeding results on all other medical datasets. Direct3D (IoU: 0.098; Dice: 0.174) and Hi3DGen (IoU: 0.120; Dice: 0.199) also performed relatively well on Duke C-Spine, confirming it as the most amenable medical dataset for single-slice reconstruction. Duke C-Spine similarly achieved the best distance metrics among medical datasets, with SAM3D attaining the lowest CD (0.153) and EMD (0.177) of any medical dataset result across both views, followed by Direct3D (CD: 0.236; EMD: 0.254). At the other extreme, MSD Lung proved the most challenging dataset: even the best-performing model (SAM3D) achieved only CD of 0.366 and EMD of 0.348, while TripoSG produced the worst overall scores (CD: 0.622; EMD: 0.607) and extremely low voxel overlap (IoU: 0.014; Dice: 0.027).

\begin{figure}[!t]
    \centering
    \makebox[\linewidth][c]{%
    \includegraphics[width=1.4\textwidth]{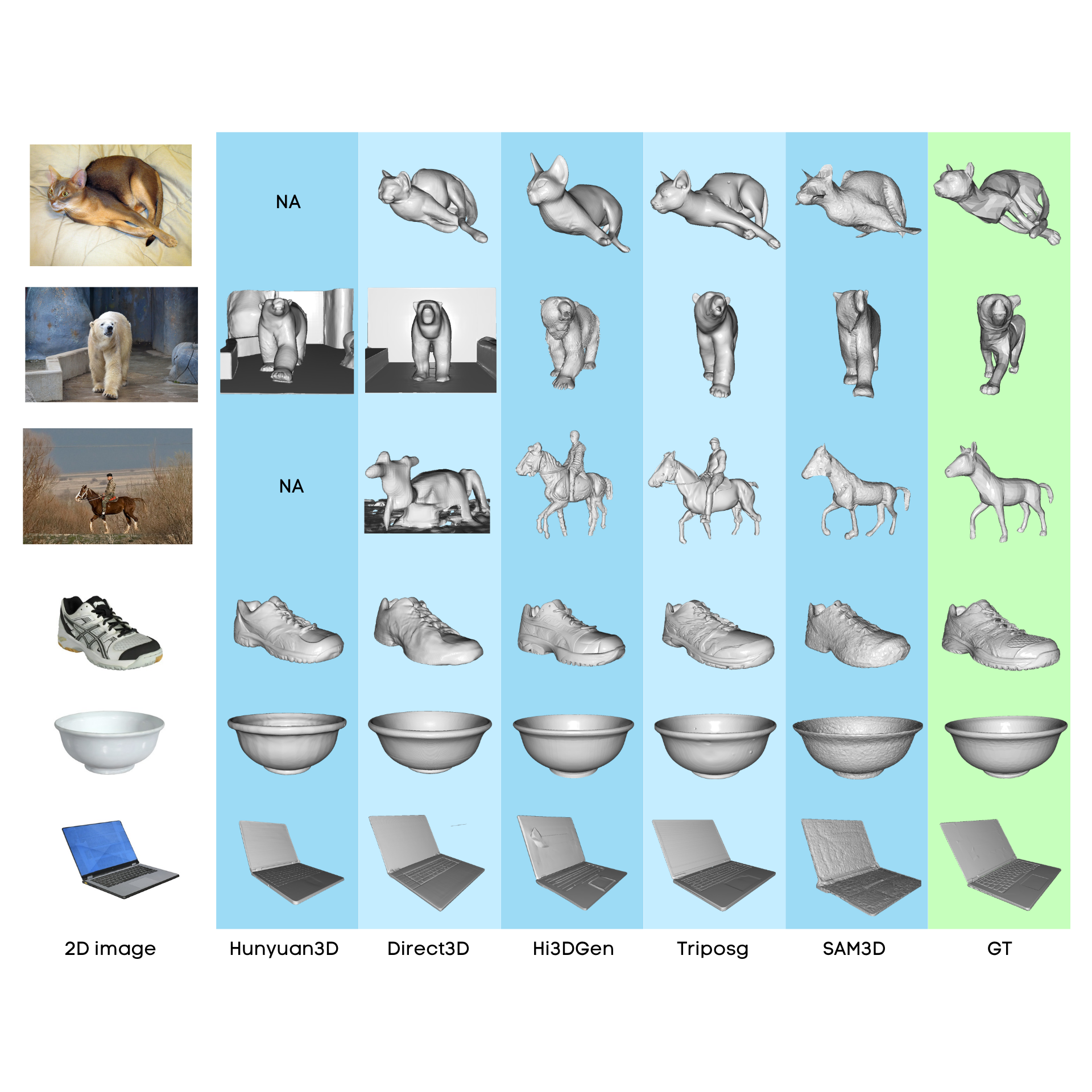}
    }
    \caption{
    \textbf{Qualitative reconstruction results on natural images.} Each row shows the 2D input, reconstructions from five models (Hunyuan3D, Direct3D, Hi3DGen, TripoSG, SAM3D), and the ground-truth 3D shape (GT) for three animals and three everyday objects. Additional examples are provided in Appendix~\ref{secC}.
    }
    \label{fig:qualitative_results_natural}
\end{figure}

\begin{figure}[!htbp]
    \centering
    \makebox[\linewidth][c]{%
    \includegraphics[width=1.05\textwidth]{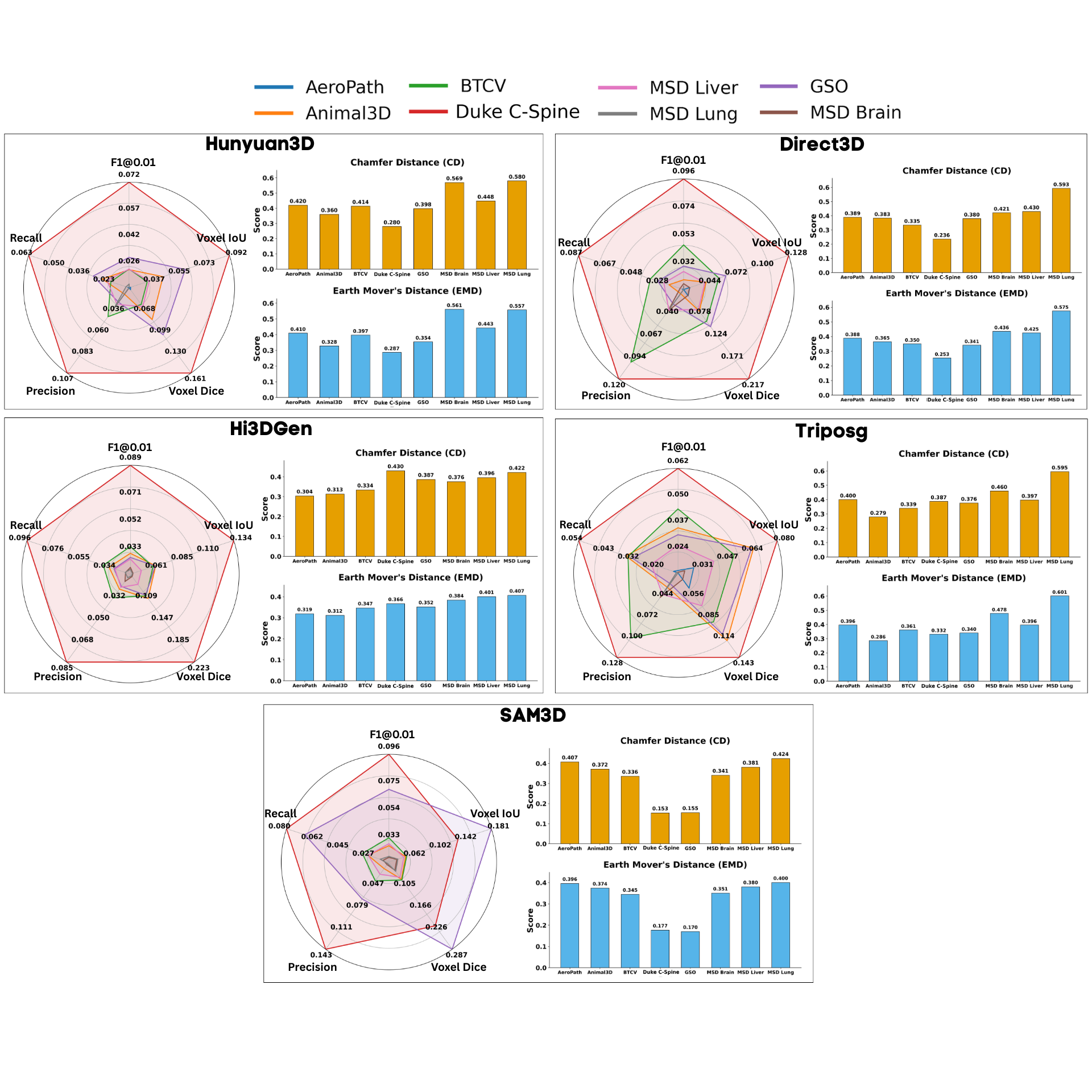}
    }
    \caption{
    \textbf{Cross-dataset performance summary for coronal reconstructions.} (a)~Spider charts of dataset-wise average F1, Voxel IoU, Voxel Dice, Precision, and Recall for each model. (b)~Average Chamfer Distance and EMD per dataset across all five models.
    }
    \label{fig:dataset_profile}
\end{figure}

\begin{figure}[!htbp]
    \centering
    \makebox[\linewidth][c]{%
    \includegraphics[width=1.05\textwidth]{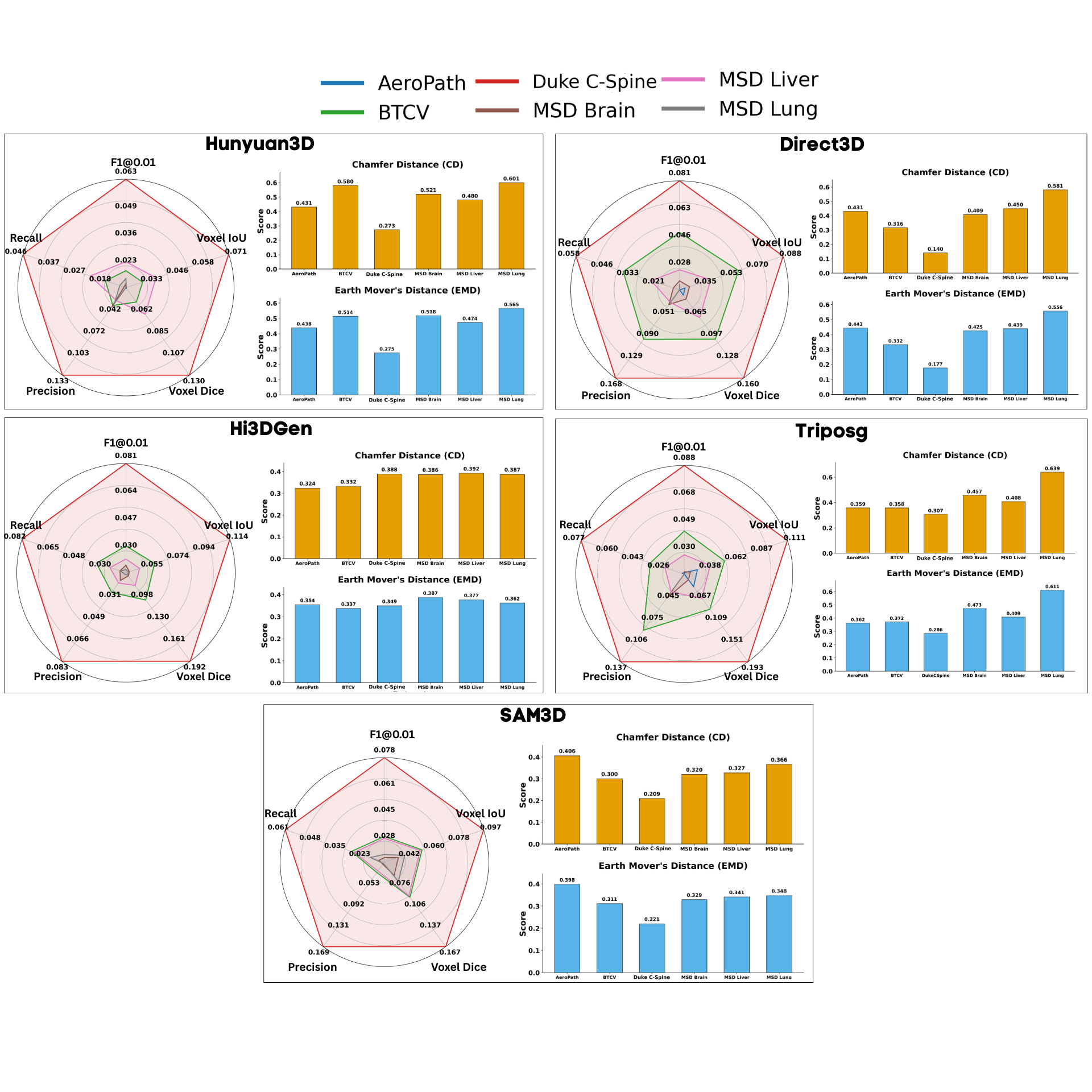}
    }
    \caption{
    \textbf{Cross-dataset performance summary for axial reconstructions.} (a)~Spider charts of dataset-wise average F1, Voxel IoU, Voxel Dice, Precision, and Recall for each model. (b)~Average Chamfer Distance and EMD per dataset across all five models.
    }
    \label{fig:dataset_profile}
\end{figure}

%% file: sec/discussion.tex
\section{Discussion}\label{sec:discussion}

\subsection{Fundamental Limits of Single-Slice 3D Reconstruction}
The consistently low voxel-based metrics across all models underscore a fundamental challenge in medical single-slice-to-3D reconstruction: 2D medical slices lack the shading gradients, occlusion boundaries, and multi-object depth cues that these foundation models were trained to exploit in natural images \cite{li2025cue3d,danier2025depthcues}, leaving depth inference severely underconstrained \cite{saxena2009make3d}. The observed discrepancies between coronal and axial reconstructions further suggest that the informativeness of the input silhouette, which is shaped by the selected anatomical plane, directly modulates reconstruction fidelity \cite{henderson2020learning}. The effect is illustrated in Figure~\ref{fig:coronal_vs_axial}, where the same MSD Brain tumor produces markedly different reconstructions depending on the input plane. This implies that plane selection is itself a non-trivial design decision for any single-view medical reconstruction pipeline. The weaker performance on pathological structures relative to anatomical ones likely reflects the greater morphological irregularity and non-convexity of tumors \cite{liu2022deep}, which deviate further from the compact, smooth shape priors these models have internalized from natural object distributions \cite{wu2018shapehd}. Together, these observations motivate both multi-view reconstruction strategies to resolve depth ambiguity \cite{wu2024multi} and domain-specific adaptation to better capture the geometric complexity characteristic of pathological anatomy \cite{guan2021domain,cui2024surgical}.

The modest advantage of SAM3D and Hi3DGen on voxel metrics, paired with the near-complete volumetric failure of TripoSG and Hunyuan3D-2.1 on certain datasets, can be understood through a shared failure mode: depth reconstruction collapse. Because voxel-based metrics are highly sensitive to depth errors \cite{tatarchenko2019single}, and medical slices lack traditional depth cues, all methods tend to produce near-planar reconstructions with minimal volumetric extent, as visually evident in the depth renderings of Figure~\ref{fig:qualitative_results}. Under this regime, even small differences in a model's ability to extrude geometry beyond the input plane translate into disproportionate gains in voxel overlap, explaining the narrow and inconsistent separation between leading and trailing methods. The low absolute scores across all models therefore reflect less a ranking of model capability than a fundamental domain mismatch: the geometric priors learned from natural objects do not transfer to the planar, texturally uniform nature of medical slices, establishing a ceiling on single-view voxel performance that model architecture alone is unlikely to overcome.

\subsection{Distance Metrics Reveal Global Shape Fidelity Differences}
The discriminative power of distance metrics over voxel metrics reveals an important nuance: while all models fail to recover accurate local volumetric structure due to depth ambiguity, they differ meaningfully in preserving global shape topology. SAM3D's consistent advantage on CD and EMD suggests that it better captures the overall point cloud distribution and coarse morphological features, even when fine-grained depth estimation remains poor. The persistent performance gap between coronal and axial inputs on distance metrics further reinforces that plane selection governs not only local voxel overlap but also global shape fidelity, as different cross-sections expose varying degrees of structural information to the model \cite{yang2025comparative} (see Figure~\ref{fig:coronal_vs_axial}). The elevated CD and EMD on pathological datasets relative to anatomical ones indicate that irregular, non-convex tumor morphologies pose a compounded challenge: they deviate further from the smooth shape priors learned from natural objects, degrading reconstruction quality at both local and global scales. These findings suggest that while SAM3D is the strongest current foundation model for zero-shot medical reconstruction, reliable recovery of pathological structures will likely require either multi-view input strategies or domain-adapted priors that explicitly encode anatomical and pathological variability.

\subsection{Domain Alignment (Medical vs. Natural) Drives Performance Differences}

The substantially stronger performance on natural datasets confirms that these foundation models operate most effectively within their training distribution, where input images contain the shading, texture, and occlusion cues necessary for reliable depth inference. SAM3D's clear dominance on GSO, a dataset of common household objects, likely reflects a strong alignment between its learned priors and the compact, well-defined geometries typical of manufactured items. The shift in competitive ranking on Animal3D, where TripoSG and Hi3DGen outperform SAM3D, suggests no single model holds a universal advantage across all object categories. Animal reconstruction may demand different geometric priors, such as articulated pose estimation and non-rigid body modeling, which other architectures capture more effectively. The quantifiable domain gap between natural and medical distance metrics (roughly 0.15 to 0.20 in absolute CD and EMD) provides a concrete measure of how much reconstruction quality degrades outside the training domain. Crucially, this establishes a performance baseline that domain-specific adaptation or multi-view strategies must overcome to make foundation model-based medical 3D reconstruction clinically viable.

\subsection{Pathological Structures Pose Compounded Reconstruction Challenges}

The disproportionately weak performance on pathological datasets highlights a compounded difficulty in reconstructing tumor and lesion structures from a single slice \cite{liu2023deep}. Whole-organ anatomical structures tend to exhibit smoother, more convex geometries that partially align with the shape priors learned from natural objects. In contrast, pathological structures feature highly irregular boundaries, non-convex morphologies, and fine-scale surface details that deviate substantially from any distribution these models encountered during training, as illustrated by the tumor reconstructions in Figure~\ref{fig:qualitative_results}. This structural complexity not only exacerbates the depth ambiguity inherent to single-slice 3D reconstruction but also challenges the models' ability to recover even a coarse global shape approximation, as reflected in the elevated CD and EMD values. Accurate characterization of tumor morphology is central to clinical tasks such as staging, surgical planning, and treatment response monitoring \cite{fang2020consensus,eisenhauer2009new}. Therefore, these findings underscore the need for targeted strategies beyond zero-shot inference, such as domain-adapted training on pathological data, multi-view input aggregation, or hybrid approaches combining foundation model priors with anatomically informed constraints.

\subsection{Geometric Complexity Drives Reconstruction Difficulty}
The pronounced performance gap between Duke C-Spine and MSD Lung highlights that geometric complexity of the target structure, rather than imaging modality alone, is a primary determinant of reconstruction difficulty. The spinal cord's simple, elongated geometry provides a more constrained target that partially aligns with the compact shape priors learned from natural objects, a similarity reflected in Duke C-Spine's distance metrics approaching those of the natural GSO and Animal3D datasets. Conversely, the irregular, non-convex morphologies of lung tumors in MSD Lung deviate maximally from these learned priors, compounding the depth ambiguity inherent to single-slice inputs and resulting in near-complete volumetric reconstruction failure for several models. The consistent advantage of natural datasets on distance metrics reaffirms that these models operate best within their training distribution, relying on traditional visual cues for depth inference. Together, these findings suggest prioritizing domain adaptation for anatomically complex and pathologically irregular structures, while geometrically simpler structures may already benefit meaningfully from zero-shot inference.

%% file: sec/method.tex
\section{Method}\label{sec:method}
\subsection{Experimental Design}


We designed a zero-shot evaluation framework to assess whether geometric priors learned by image-to-3D foundation models from natural images could transfer to medical imaging data. The overall pipeline, illustrated in Figure 1, proceeded in four stages: slice extraction, masking, 3D reconstruction, and quantitative evaluation.

For each volumetric sample in a given dataset, we selected a midpoint slice along either the coronal or axial anatomical plane from the NIfTI scan volume. We extracted the corresponding 2D binary segmentation mask at the same slice index from the paired NIfTI mask volume. The binary mask was then applied to the 2D image slice to isolate the structure of interest, producing a single masked 2D input image $I \in \mathbb{R}^{H \times W}$ that served as the sole input to each reconstruction model. To construct the ground-truth 3D representation, we extracted surface points from the full 3D segmentation mask using morphological erosion: boundary voxels were identified as the set difference between the original mask and its eroded counterpart, and the resulting coordinates were scaled by voxel spacing to yield a point cloud $P \in \mathbb{R}^{N \times 3}$ in physical units.

\subsection{Datasets}

\begin{table}[t]
\makebox[\linewidth][c]{%
\centering

\begin{tabular}{p{0.18\linewidth} p{0.12\linewidth}  p{0.10\linewidth} p{0.45\linewidth} p{0.10\linewidth}}
\toprule
\textbf{Dataset} & \textbf{Type} & \textbf{Modality} & \textbf{Structures Present }  & \textbf{Samples} \\
\midrule
AeroPath \cite{stoverud2023aeropath} & Anatomical & CT & Lungs, Airways & 54 \\
BTCV \cite{landman2015btcv} & Anatomical & CT  & Spleen, Kidneys, Gall Bladder, Esophagus, liver heart, pancreas, adrenal gland & 388 \\
Duke C-Spine \cite{zhou2025cspineseg} & Anatomical & MRI  & Spinal Cord & 2510 \\
MSD Lung \cite{antonelli2022msd} & Pathological & CT & Lungs & 63 \\
MSD Brain \cite{antonelli2022msd} & Pathological & MRI & Brain & 1440 \\
MSD Liver \cite{antonelli2022msd} & Pathological & CT  & Liver & 249 \\
GSO \cite{downs2022scannedobjects} & Natural & Camera & Common Household Items & 1030 \\
Animal3D \cite{xu2023iccv} & Natural  & Camera & 40 Mammal Species & 3379 \\

\bottomrule
\end{tabular}
}
\caption{
\textbf{Datasets used in this study.} Six medical datasets span anatomical and pathological structures across CT and MRI; two natural datasets provide camera-captured objects for comparison. Sample counts for medical datasets denote unique 3D reconstructions per plane; the total is doubled as both coronal and axial views are evaluated.
}
\label{tab:datasets_used}
\end{table}

We evaluated reconstruction performance across eight datasets spanning medical and natural domains, summarized in Table~\ref{tab:datasets_used}. The medical datasets comprised six collections of volumetric NIfTI (.nii) files, each containing a 3D scan paired with a corresponding segmentation mask. The natural datasets provided camera-captured images of real-world objects with associated ground-truth 3D geometry. This selection was designed to probe generalization across imaging modalities (CT and MRI), structure types (anatomical and pathological), and domain boundaries (medical and natural).

\noindent\textbf{Medical Datasets.}
The six medical datasets were divided into anatomical and pathological categories. The anatomical datasets (AeroPath~\cite{stoverud2023aeropath}, BTCV~\cite{landman2015btcv}, and Duke C-Spine~\cite{zhou2025cspineseg}) contained segmentations of whole-organ or whole-structure anatomy, including lungs, airways, abdominal organs, and spinal cord. The pathological datasets (MSD Lung, MSD Brain, and MSD Liver~\cite{antonelli2022msd}) contained segmentations of tumors within their respective organs. This distinction was clinically motivated: anatomical structures tend to exhibit smoother, more convex geometries, whereas pathological structures such as tumors present highly irregular and non-convex morphologies that pose a greater reconstruction challenge. 

\noindent\textbf{Natural Datasets.}
Two natural datasets served as in-distribution baselines for the foundation models. Google Scanned Objects (GSO)~\cite{downs2022scannedobjects} comprised 1,030 high-fidelity 3D scans of common household items, representing the compact, well-defined geometries typical of manufactured objects. Animal3D~\cite{xu2023iccv} contained 3,379 samples across 40 mammal species, capturing articulated and non-rigid body geometries that required different shape priors than manufactured items. These datasets allowed us to quantify the domain gap between medical and natural reconstruction performance.

\noindent\textbf{Medical Data Preprocessing.}
Each medical instance consisted of two NIfTI files: a volumetric scan and its corresponding binary segmentation mask (Figure~\ref{fig:SAM3D_fig}B). The scan stored a 3D anatomical or pathological structure as a stack of 2D slices along three orthogonal planes (sagittal, coronal, and axial). To construct the ground-truth 3D surface, we extracted boundary voxels from the segmentation mask via morphological erosion, computed as the set difference between the original mask and its eroded version, and scaled the resulting coordinates by voxel spacing to obtain a point cloud in physical units. For the 2D input, we selected the midpoint slice along the chosen anatomical plane and extracted both the image and binary mask at that slice index. The mask was applied to the image to produce the final masked 2D input (Figure~\ref{fig:SAM3D_fig}C). This yielded, for each sample, a masked 2D image $I \in \mathbb{R}^{H \times W}$ and a ground-truth 3D point cloud $P \in \mathbb{R}^{N \times 3}$.

\noindent\textbf{Plane Selection.}
We evaluated reconstructions along the coronal and axial planes only (Figure~\ref{fig:SAM3D_fig}A). Sagittal views were excluded as they provided less consistently informative cross-sections of the target structures and introduced additional variability without corresponding diagnostic benefit.
Natural datasets (GSO and Animal3D) were evaluated using their native single-view input images without plane selection, as they were captured by cameras rather than volumetric scanners.

\subsection{Model Comparisons} \label{model_comparisons}

We benchmarked SAM3D~\cite{sam3dteam2025sam3d3dfyimages} against four state-of-the-art diffusion-based image-to-3D models that collectively span the dominant reconstruction paradigms. Hunyuan3D-2.1~\cite{hunyuan3d2025hunyuan3d21} combined a diffusion transformer for shape generation with a diffusion-based texture module in a large-scale image-conditioned pipeline. Direct3D~\cite{wu2024direct3d} operated as a 3D-native latent diffusion transformer that generated assets directly in a compact 3D latent space. Hi3DGen~\cite{ye2025hi3dgen} targeted fine geometry generation through an intermediate normal-bridging representation. TripoSG~\cite{li2025triposg} employed large-scale rectified-flow generative modeling to synthesize 3D shapes with high geometric fidelity. All five models were evaluated using their default inference configurations in a zero-shot setting, without any fine-tuning or domain-specific adaptation.

\subsection{Metrics} \label{metrics}

Following the evaluation protocol in \cite{sam3dteam2025sam3d3dfyimages}, we normalized both the predicted and ground-truth point clouds to a unit cube $[-1, 1]^3$ by centering each at its centroid and scaling by the maximum absolute coordinate, then applied point-to-point Iterative Closest Point (ICP) registration with a correspondence threshold of $0.02$ to align the prediction to the ground truth. We assessed reconstruction quality using five metrics. F1@0.01 measured surface accuracy and coverage at a distance threshold $\tau = 0.01$: for each predicted point we queried its nearest neighbor in the ground truth via a $k$-d tree and defined precision as the fraction of predicted points within $\tau$, with recall defined symmetrically; F1 was the harmonic mean of the two. Voxel IoU and Voxel Dice both discretized the aligned point clouds onto a $64^3$ grid spanning $[-1, 1]^3$, assigning each point to a voxel by floor-dividing its coordinates by the voxel size and clipping to grid bounds; Voxel IoU was computed as $|V_{\hat{P}} \cap V_P| / |V_{\hat{P}} \cup V_P|$ and Voxel Dice as $2|V_{\hat{P}} \cap V_P| / (|V_{\hat{P}}| + |V_P|)$, where $V_{\hat{P}}$ and $V_P$ denoted the occupied voxel sets for the predicted and ground-truth shapes, respectively. Chamfer Distance (CD) summed the mean nearest-neighbor $\ell_2$ distances in both directions using $k$-d tree queries: $\text{CD} = \frac{1}{|\hat{P}|}\sum_{\hat{p}}\min_{p}\|\hat{p}-p\|_2 + \frac{1}{|P|}\sum_{p}\min_{\hat{p}}\|p-\hat{p}\|_2$. Earth Mover's Distance (EMD) first subsampled both point clouds to at most 2{,}048 points, then solved the optimal bijection via the Hungarian algorithm on the pairwise Euclidean cost matrix and reported the mean assignment cost. For F1, Voxel IoU, and Voxel Dice, higher values indicated better reconstruction; for CD and EMD, lower values indicated better reconstruction.